\pdfoutput=1

\documentclass[11pt]{article}

\usepackage{ACL2023}

\usepackage{times}
\usepackage{latexsym}

\usepackage[T1]{fontenc}

\usepackage[utf8]{inputenc}

\usepackage{microtype}

\usepackage{tabularx}
\usepackage{enumitem}
\usepackage{inconsolata}
\usepackage{graphicx}
\usepackage{bm}
\usepackage{bbm}
\usepackage{multirow}
\usepackage{xcolor}
\usepackage{colortbl}
\usepackage{makecell}

\title{OPI at SemEval 2023 Task 9: A Simple But Effective Approach to Multilingual Tweet Intimacy Analysis}

\author{Sławomir Dadas\\
  National Information Processing Institute, Warsaw, Poland\\
  \texttt{sdadas@opi.org.pl}}

\begin{document}
\maketitle
\begin{abstract}
This paper describes our submission to the SemEval 2023 multilingual tweet intimacy analysis shared task. The goal of the task was to assess the level of intimacy of Twitter posts in ten languages. The proposed approach consists of several steps. First, we perform in-domain pre-training to create a language model adapted to Twitter data. In the next step, we train an ensemble of regression models to expand the training set with pseudo-labeled examples. The extended dataset is used to train the final solution. Our method was ranked first in five out of ten language subtasks, obtaining the highest average score across all languages. 
\end{abstract}

\section{Introduction}

Intimacy can be expressed in language in a variety of ways. The degree of intimacy in an utterance is indicated by both thematic and stylistic features, often subtle and difficult to quantify automatically. One of the most apparent aspects of intimacy is self-disclosure. Sharing personal details about oneself or one's life can create a sense of intimacy. This information can relate to both factual as well as emotional spheres, addressing matters such as feelings, goals, dreams, or fears. Other features which may indicate intimacy involve the use of certain types of terms or phrases, especially those creating a sense of closeness and connection between the author and the reader.

Automatic identification and quantification of intimacy in natural language is a challenging problem, with a difficulty similar to automatic emotion recognition. Both tasks involve measuring inherently subjective and ambiguous aspects. Intimacy can be influenced by a variety of factors such as the context, culture, and personal experiences of the individual. Intimacy analysis in multilingual text presents additional challenges due to differences in language structure, cultural norms, and expression of emotions across different languages. So far, this topic has not received much attention. \citet{pei2020quantifying} conducted an intimacy analysis of questions from social media, books, and movies. In the study, they created a question dataset and examined the performance of automatic intimacy prediction using methods such as logistic regression and transformer-based language models \citep{Devlin2019bert,liu2019roberta}.

The multilingual intimacy analysis task was a part of SemEval 2023. The goal of the task was to measure the intimacy of Twitter posts in ten languages \citep{pei2022semeval}. The organizers provided a training set of 9,491 tweets and a test set of 13,797 tweets, in which each sample was annotated with an intimacy score ranging from 1 to 5. The training data included texts in only six of the ten languages, while the evaluation was performed on all ten. The task thus tested the performance of the submitted solutions for both standard fine-tuning and zero-shot prediction. The metric selected to evaluate the solutions was the Pearson correlation coefficient. The systems were ranked according to the correlation value for the entire test set, as well as for subsets in each language.

This paper presents our solution to the multilingual tweet intimacy analysis shared task. The proposed approach is a combination of domain adaptation and semi-supervised learning. We first train a transformer language model on a large corpus of multilingual tweets and then create an ensemble of regression models to expand the training set with pseudo-labeled examples. Our method achieved the best score in five out of ten language subtasks, the highest number among all participants. According to the Pearson correlation calculated for the entire dataset, the proposed method was ranked third.

\section{System description}
\begin{figure*}
  \centering
  \includegraphics[scale=0.6]{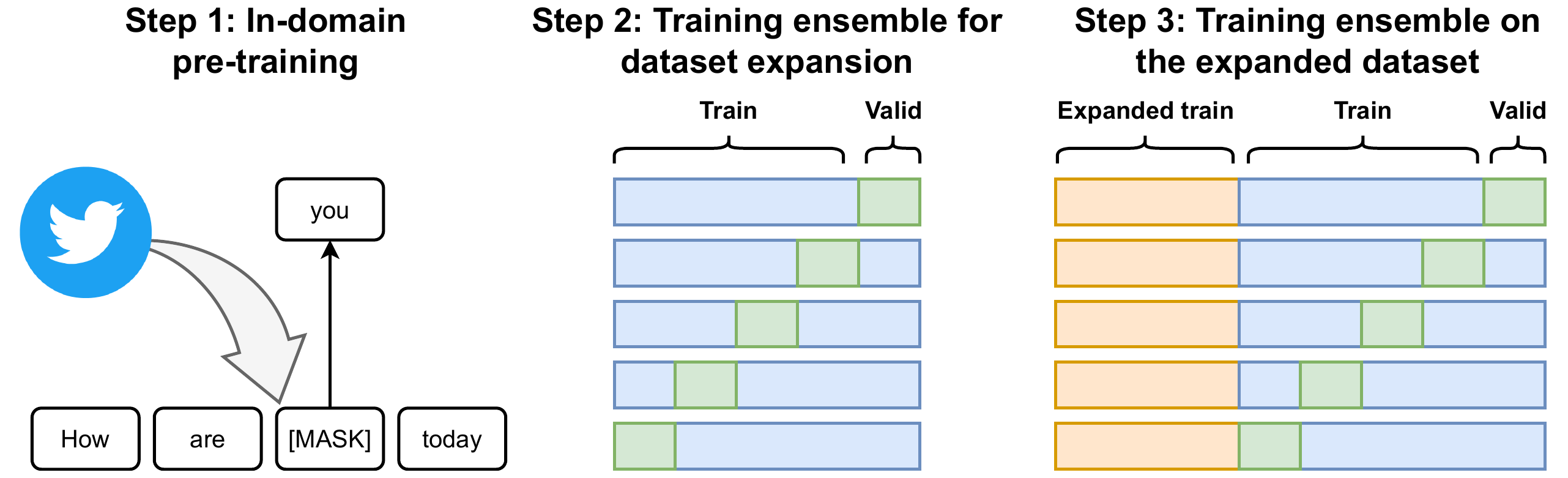}
  \caption{Our solution to the multilingual tweet intimacy analysis task. First, we further pre-train an existing language model on a large corpus of tweets. Next, we train an ensemble, which is used to label additional data. The expanded dataset is utilized to create a new set of models, which are the final solution to the task.}
  \label{fig:model}
\end{figure*}

Our solution for the multilingual tweet intimacy analysis task can be summarized in the following three steps:
\begin{enumerate}[wide,labelwidth=0pt,labelindent=0pt,itemsep=0pt,topsep=8pt]
\item We adapt a transformer language model to the problem domain by fine-tuning it on a large corpus of tweets in multiple languages. The model trained by us has been made publicly available.\footnote{\url{https://huggingface.co/sdadas/xlm-roberta-large-twitter}}
\item We employ the fine-tuned language model to train an ensemble of regressors. These models are then used to label new data, which we append to the original training set.
\item We train a new ensemble on the expanded training set, which is used to generate the final predictions.
\end{enumerate}
Figure \ref{fig:model} shows our approach on a diagram. In the following sections, we explain the steps of this process in detail.

\subsection{Domain adaptation}

The importance of adapting language models to domains and tasks has been highlighted in the scientific literature in recent years \citep{howard2018universal,gururangan2020don,ramponi2020neural}. By further pre-training the language model on data from the target domain, the model can better capture the language patterns and nuances specific to that domain, resulting in improved accuracy and performance. Additionally, by leveraging knowledge and patterns learned from the source domain, the language model can be trained more effectively on a supervised task, needing fewer data samples. In the case of Twitter data, this is particularly relevant, as the community on this platform uses a specific language, which differs from the typical texts on which publicly available language models have been trained. It is characterized by the use of non-standard abbreviations, acronyms, and truncated words, making the language more informal and less structured. It is also common practice to replace words or phrases with hashtags. Additionally, due to the limited character count and informal nature of Twitter, users may not always adhere to traditional spelling and grammar rules. Common deviations include the omission of articles and prepositions, the use of contractions and slang, and the omission of punctuation.

\begin{table}[h]
\small
\centering
\setlength{\tabcolsep}{4pt}
\renewcommand{\arraystretch}{1.3}
\begin{tabular}{l|p{1cm}|p{1cm}|p{1cm}|p{1cm}}
\hline
\textbf{Language} & \multicolumn{2}{c|}{\thead{\textbf{In-domain} \\ \textbf{pre-training data}}} & \multicolumn{2}{c}{\thead{\textbf{Pseudo-labeled} \\ \textbf{training data}}} \\
\hline
English (EN) & 79.4m & 50.9\% & 37.2t & 12.7\%\\
Spanish (ES) & 22.4m & 14.4\% & 49.7t & 17.0\%\\
Portuguese (PT) & 16.3m & 10.5\% & 42.9t & 14.7\%\\
Italian (IT) & 2.5m & 1.6\% & 37.6t & 12.9\%\\
French (FR) & 6.6m & 4.2\% & 34.6t & 11.8\%\\
Chinese (ZH) & 4.0m & 2.5\% & 25.4t & 8.7\%\\
Hindi (HI) & 2.7m & 1.7\% & 23.9t & 8.2\%\\
Dutch (NL) & 1.1m & 0.7\% & 17.5t & 6.0\%\\
Korean (KO) & 8.3m & 5.3\% & 12.6t & 4.3\%\\
Arabic (AR) & 12.9m & 8.2\% & 11.2t & 3.8\% \\
\hline
\textbf{Total} & 156.2m & 100\% & 292.5t & 100\% \\
\hline
\end{tabular}
\caption{\label{tab:dataset}
Distribution of tweets by language in the pre-training and expanded training dataset. The number of tweets (in millions or thousands) and the percentage of each language in the datasets are shown.
}
\end{table}

The basis of our solution is the XLM RoBERTa large model \citep{conneau2020unsupervised}. It is a transformer-based language model, trained on a dataset of 100 languages. In order to adapt this model to the Twitter domain, we further optimized it utilizing masked language modeling (MLM) on a dataset of over 156 million tweets. The dataset was derived from \emph{archive.org} Twitter stream collection\footnote{\url{https://archive.org/details/twitterstream}}, from which we extracted data spanning four months, from May to August 2021. Next, we discarded all posts shorter than 20 characters and written in languages other than those covered by the shared task. We also applied the same preprocessing procedure as the authors of XLM-T \citep{barbieri-etal-2022-xlm}, replacing all usernames with the string \emph{@user} and all URLs with \emph{http}\footnote{\url{https://huggingface.co/cardiffnlp/twitter-xlm-roberta-base}}. Table \ref{tab:dataset} shows the number and percentage of records from each language included in the pre-training dataset.

The model was trained for two epochs. We used a learning rate scheduler with warmup and polynomial decay. The peak learning rate was set to 2e-5 and the warmup phase lasted for 6\% iterations of the first epoch. We trained the model with a batch size of 1024 on eight Nvidia V100 graphic cards for two weeks.

\subsection{Dataset expansion}
The second stage of our solution was to expand the training set by automatically labeling additional data. For this, we employed a method known as pseudo-labeling \citep{lee2013pseudo}. It is a semi-supervised learning technique in which a model is first trained on a small set of labeled data, and then used to predict the labels of the remaining unlabeled data. These predicted labels are then added to the training set as if they were actual labels, creating a larger dataset that can be used to retrain the model.

In our approach, an ensemble of five regression models was used to predict the scores for unlabeled examples. The procedure we used involved dividing the original training set into five equal parts. This created five possible data splits, with each split consisting of 80\% training data and 20\% intended for validation. For each such split, we trained five regression models with different random seeds and then selected the model achieving the highest Pearson correlation value on the validation part. The process thus consisted of training a total of 25 models (5 splits, 5 models per split), from which the best five were selected, one for each split. From these models, an ensemble was created. The individual models were fine-tuned with MSE loss and a batch size of 32 for three epochs. A learning rate scheduler with warmup and polynomial decay was used with a peak learning rate of 1e-5.

The trained ensemble was used for pseudo-labeling. For each sample from a corpus of 156 million tweets, we calculated the intimacy score as the mean value of the predictions returned by the models. In addition, we also calculated the standard deviation of each score. Our intention was to include only the samples in the expanded dataset, which were predicted with high confidence. Accordingly, we set the threshold at 0.05 and only tweets with a standard deviation below this value were selected. In order to create a more balanced dataset, we also imposed additional limits on the number of records with similar characteristics. The number of examples from the same language and having the same range of intimacy scores (e.g. from 2.0 to 3.0) could not exceed 10 thousand. Using the described procedure, we were able to extract the dataset of over 292 thousand pseudo-labeled examples, the distribution of which is shown in Table \ref{tab:dataset}. 

\subsection{Generating predictions}
In the last step, we add pseudo-labeled examples to the training dataset and create a new ensemble of regressors, which was used to generate the final results. The procedure for training the models is similar to the one previously described. Once again, we split the original dataset into five parts, one part of which we use for model validation. In this case, however, for training in addition to the other four parts of the original data, we also used the entire pseudo-labeled dataset. As before, we trained 25 models and selected the best one from each split to form the final ensemble. Predictions for the test dataset were calculated as the mean value of the outputs from the individual models.

\begin{table*}
\small
\centering
\setlength{\tabcolsep}{5pt}
\renewcommand{\arraystretch}{1.3}
\begin{tabular}{l|c|c|cccccccccc}
\hline
\textbf{System} & \textbf{ALL} & \textbf{AVG} & \textbf{EN} & \textbf{ES} & \textbf{PT} & \textbf{IT} & \textbf{FR} & \textbf{ZH} & \textbf{HI} & \textbf{NL} & \textbf{KO} & \textbf{AR} \\
\hline
Ohio State University & \color{blue} \textbf{0.616} & 0.635 & 0.758 & 0.770 & 0.689 & \color{purple} \textbf{0.739} & \color{purple} \textbf{0.726} & \color{purple} \textbf{0.756} & 0.226 & 0.623 & \color{purple} \textbf{0.414} & 0.643 \\
University of Zurich & \color{purple} \textbf{0.614} & 0.616 & 0.722 & 0.740 & 0.689 & 0.723 & 0.710 & 0.718 & 0.224 & 0.619 & 0.380 & 0.636 \\
\rowcolor[HTML]{eaecf0} \textbf{Our system} & 0.613 & \color{blue} \textbf{0.638} & 0.749 & \color{purple} \textbf{0.775} & \color{blue} \textbf{0.702} & \color{blue} \textbf{0.743} & 0.695 & \color{blue} \textbf{0.763} & 0.238 & \color{blue} \textbf{0.679} & 0.370 & \color{blue} \textbf{0.663} \\
University of Tyumen & 0.599 & 0.621 & 0.717 & 0.740 & 0.684 & 0.734 & 0.708 & 0.721 & 0.242 & 0.639 & 0.361 & \color{purple} \textbf{0.662} \\
NetEase Inc & 0.599 & 0.619 & 0.728 & 0.746 & 0.699 & 0.735 & 0.701 & 0.734 & 0.223 & 0.640 & 0.333 & 0.652 \\
\hline
\end{tabular}
\caption{\label{tab:results_main}
The performance of five top-rated teams in the multilingual tweet intimacy analysis task according to the official results. We show the Pearson correlation value for the entire dataset (ALL), for individual language subtasks, and the average correlation value across all languages (AVG). Blue color indicates the best score in a given category among all participants, red color indicates the second-best score.
}
\end{table*}

\section{Experiments and results}
This section contains a discussion of the official results of the multilingual tweet intimacy analysis task. We also conducted post-evaluation experiments using the gold labels provided by the organizers to analyze the results obtained by models different from the submitted solution.

\subsection{Official results}
The evaluation covered ten languages, six of which were present in the training data, whereas four appeared only in the test dataset. The set of seen languages included English, Spanish, Italian, Portuguese, French, and Chinese. The set of unseen languages, intended to test the performance of solutions in a zero-shot setting, included Hindi, Arabic, Dutch, and Korean. 45 teams participated in the shared task. Our solution was ranked third in the main classification. Our method scored high on all but two languages. The weaker points of our solution were French and Korean, on which we were ranked 13th and 12th, respectively. We won in five language-specific subtasks and placed second in one. We also obtained the highest average correlation value on all languages among the submitted solutions. The results of the top five ranked solutions according to the correlation value for the entire test set are shown in Table \ref{tab:results_main}.

Based on the results, we can observe a problem associated with the Pearson correlation coefficient, which was chosen as an evaluation metric. In the general case, the value of this coefficient for disjoint subgroups of the population may not necessarily be related to the value for the entire population. In the case of the discussed task, the results for individual languages are not fully aligned with the results on the entire test set. This can be examined by comparing the coefficient value for the whole dataset and the average value of the coefficient for all languages. In the case of the latter value, among the top participants, only our solution and the winning solution achieved high performance across languages. Although there were other participants who obtained an average value above 0.63, they were ranked lower, even outside of the top ten teams. For example, the 12th-placed team achieved the best score in two languages and was in the top three in six. The overall Pearson coefficient for this solution was only 0.587, while the average of the coefficients was 0.636.

\subsection{Post-evaluation results}

In post-evaluation experiments, we fine-tuned publicly available multilingual language models on the tweet intimacy analysis task, comparing their results with the results obtained by the submitted solution. In the experiment, we included the original XLM RoBERTa \citep{conneau2020unsupervised} models in base and large sizes,  as well as our version of the large model tuned on a corpus of 156 million tweets. We also utilized XLM-T models, published by \citet{barbieri-etal-2022-xlm} as a part of their study on Twitter sentiment analysis. The authors trained XLM RoBERTa base model on a dataset of 198 million tweets, and then further tuned it on sentiment analysis datasets in eight languages. We evaluate both the pre-trained and fine-tuned versions of this model.

\begin{table}[h]
\small
\centering
\setlength{\tabcolsep}{5pt}
\renewcommand{\arraystretch}{1.3}
\begin{tabular}{l|cccc}
\hline
\textbf{Model} & \textbf{Avg} & \textbf{StdDev} & \textbf{Max} & \textbf{Min}\\
\hline
\multicolumn{5}{l}{\textbf{Training on the original dataset}} \\
\hline
XLM-T (pre-trained) & 0.565 & ±0.016 & 0.588 & 0.545 \\
XLM-T (sentiment) & 0.558 & ±0.022 & 0.594 & 0.530 \\
XLM-R (base) & 0.537 & ±0.008 & 0.545 & 0.522 \\
XLM-R (large) & 0.580 & ±0.016 & 0.599 & 0.561 \\
\rowcolor[HTML]{eaecf0} XLM-R (ours) & 0.602 & ±0.026 & \textbf{0.636} & 0.564 \\
\hline
\multicolumn{5}{l}{\textbf{Training on the expanded dataset}} \\
\hline
XLM-T (pre-trained) & 0.603 & ±0.001 & 0.604 & 0.602 \\
XLM-T (sentiment) & 0.598 & ±0.003 & 0.603 & 0.594 \\
XLM-R (base) & 0.590 & ±0.003 & 0.595 & 0.588 \\
XLM-R (large) & 0.595 & ±0.002 & 0.600 & 0.590 \\
\rowcolor[HTML]{eaecf0} XLM-R (ours) & 0.611 & ±0.002 & 0.614 & 0.608 \\
\hline
\multicolumn{5}{l}{\textbf{Submitted solution}} \\
\hline
\rowcolor[HTML]{eaecf0} Individual models & 0.612 & ±0.003 & 0.616 & \textbf{0.608} \\
\rowcolor[HTML]{eaecf0} Ensemble & \textbf{0.613} & - & - & - \\
\hline
\end{tabular}
\caption{\label{tab:models}
Performance comparison of the submitted solution and fine-tuned language models in two ways: on the original training data and using the extended dataset. The results shown refer to the Pearson correlation values for the entire dataset.
}
\end{table}

The comparison is shown in Table \ref{tab:models}. We demonstrate the performance of models trained only on the original training data, and those trained on the extended dataset. For each row in the table, a given model was trained five times with different random seeds. The table shows the average value of the achieved results, as well as the standard deviation, maximum and minimum values. We can see that extending the dataset with pseudo-labeled examples yielded better average results in each case, and also significantly reduced the standard deviation of the results. Training the models on the original dataset appears to be unstable, giving varying results for different runs. Interestingly, one of the fine-tuned models achieved an overall Pearson correlation of 0.636, higher than any solution in the evaluation phase. The same model scored low on individual languages, performing worse in 8 out of 10 languages compared to the solution we submitted, which once again indicates a disparity between overall and individual scores. 

The variant of XLM RoBERTa adapted by us to the Twitter domain obtained the best average correlation values for both the original and extended datasets. This shows the effectiveness of pre-training on in-domain data, as the results achieved by our model are significantly better than those of the original XLM-R models. A second choice could be XLM RoBERTa large or pre-trained XLM-T, which also achieved solid results. On the other hand, the use of the ensemble in the final submission does not seem to yield a clear improvement over the individual models. The average score obtained by the standalone models is only 0.001 lower than the ensemble solution.

\section{Conclusion}
In this paper, we described our solution for the multilingual tweet intimacy analysis shared task. Our system placed first in five out of the ten languages. The paper demonstrated a method for combining domain adaptation with semi-supervised learning. As part of our research, we trained and published a multilingual language model using a corpus of 156 million tweets. Building on this model, we fine-tuned an ensemble of regressors to extend the training dataset with pseudo-labeled examples. We also performed additional experiments, comparing the model to other publicly available multilingual models, in which our method proved to be more effective in predicting intimacy scores.

\bibliography{anthology,custom}
\bibliographystyle{acl_natbib}
\end{document}